\documentclass[11pt,a4paper]{article}
\usepackage[hyperref]{naaclhlt2018}
\usepackage{times}
\usepackage{latexsym}
\usepackage{times}
\usepackage{latexsym}
\usepackage[fleqn]{amsmath}
\usepackage{url}
\usepackage{booktabs}
\usepackage{multirow}
\usepackage{graphicx}
\usepackage{caption}
\usepackage{subcaption}
\usepackage{mathabx}
\usepackage{xparse}
\usepackage{xspace}
\usepackage{mathpartir}
\usepackage{tikz}
\usepackage{tikz-dependency}
\usepackage{etoolbox}
\usepackage{array,booktabs,ragged2e}
\usepackage{amsthm}
\usepackage{setspace}
\usepackage{bm}
\usepackage[T1,T5]{fontenc}

\DeclareTextSymbolDefault{\OHORN}{T5}
\DeclareTextSymbolDefault{\UHORN}{T5}
\DeclareTextSymbolDefault{\ohorn}{T5}
\DeclareTextSymbolDefault{\uhorn}{T5}

\aclfinalcopy %
\def\aclpaperid{948} %

\title{
Improving Coverage and Runtime Complexity for Exact Inference in Non-Projective Transition-Based Dependency Parsers}

\author{Tianze Shi \\
   Cornell University\\
  {\tt tianze@cs.cornell.edu} \\\And
  Carlos G{\'o}mez-Rodr{\'i}guez \\
  Universidade da Coru{\~n}a\\
  {\tt carlos.gomez@udc.es} \\\And
  Lillian Lee \\
  Cornell University\\
  {\tt llee@cs.cornell.edu} \\}

\date{}

\def\namecite{\newcite}

\mathchardef\mhyphen="2D %

\newcommand{\shift}{\ensuremath{\mathsf{sh}}}
\newcommand{\reduce}{\ensuremath{\mathsf{re}}}
\newcommand{\reduceargs}[2]{\ensuremath{\mathsf{re}_{#1,#2}}} %

\newcommand{\reffig}[1]{Fig.~\ref{#1}}
\newcommand{\refsec}[1]{\S\ref{#1}}

\newtheorem*{theorem*}{Theorem}

\newtheorem*{lemma*}{Lemma}

\newcommand{\terminal}{\tau}

\newcommand{\stack}{\ensuremath{\sigma}}
\newcommand{\buffer}{\ensuremath{\beta}}
\newcommand{\arcset}{\ensuremath{A}}

\newcolumntype{R}[1]{>{\RaggedLeft\arraybackslash}p{#1}}

\newcommand{\posscite}[1]{\citeauthor{#1}'s \citeyearpar{#1}} %
\newcommand{\posscitelong}[1]{\citeauthor*{#1}'s \citeyearpar{#1}}

\newcommand{\concept}[1]{{\em #1}}

\newcommand{\ruleset}{\ensuremath{{\cal R}}}

\newcommand{\varone}{\textsc{AllDeg1}\xspace}
\newcommand{\vartwo}{\textsc{All}\xspace}
\newcommand{\varthree}{\textsc{All${s_0s_1}$}\xspace}

\begin{document}
\maketitle

\begin{abstract}
We generalize
\posscitelong{attardi-dp} parser to
a family of non-projective transition-based dependency parsers
allowing polynomial-time exact inference.
This includes novel parsers with
better
coverage than \namecite{attardi-dp}, and even a variant that reduces time complexity
to $O(n^6)$, improving
on prior bounds.
We hope that this
piece of theoretical work
inspires
design of novel transition systems with better coverage
and
better
run-time guarantees.
\end{abstract}

\section{Introduction}
\label{sec:intro}

\begin{figure*}[!ht]
\centering
\begin{tabular}{ccccc}
\toprule
System & Reduce Transitions & \begin{tabular}{@{}c@{}}Non-proj. \\ Coverage\end{tabular} & \begin{tabular}{@{}c@{}} Time \\ Complexity \end{tabular} & \begin{tabular}{@{}c@{}}Max. \\ Degree\end{tabular}\\
\midrule
\namecite{attardi} &
\raisebox{-4pt}{
\includegraphics[width=0.4\textwidth]{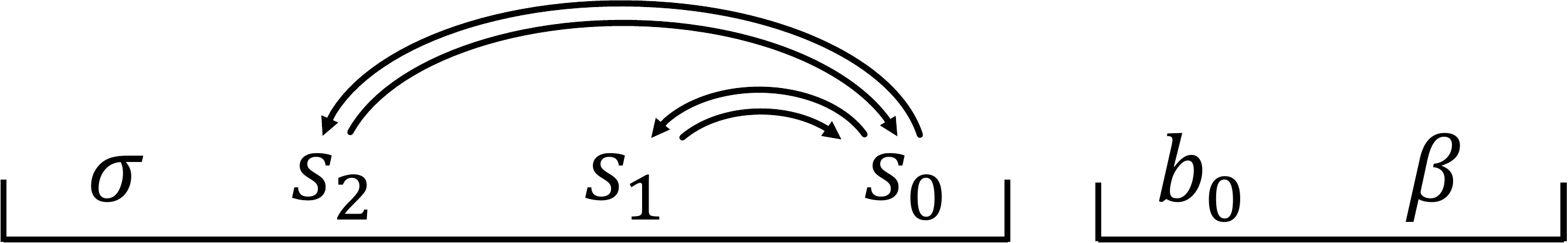}}
& $87.24\%$ & $O(n^7)$ & $2$ \\[3pt]
Var. 1: \varone &
\raisebox{-4pt}{
\includegraphics[width=0.4\textwidth]{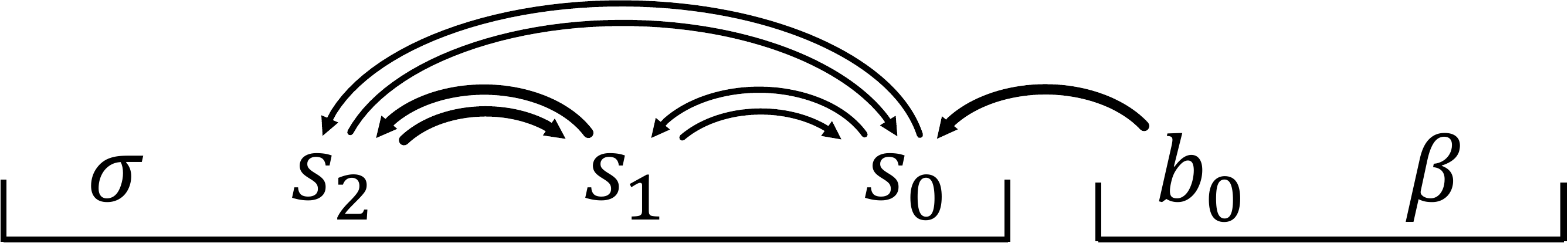}}
& $93.32\%$ & $O(n^7)$ & $2$ \\[3pt]
Var. 2: \vartwo &
\raisebox{-4pt}{
\includegraphics[width=0.4\textwidth]{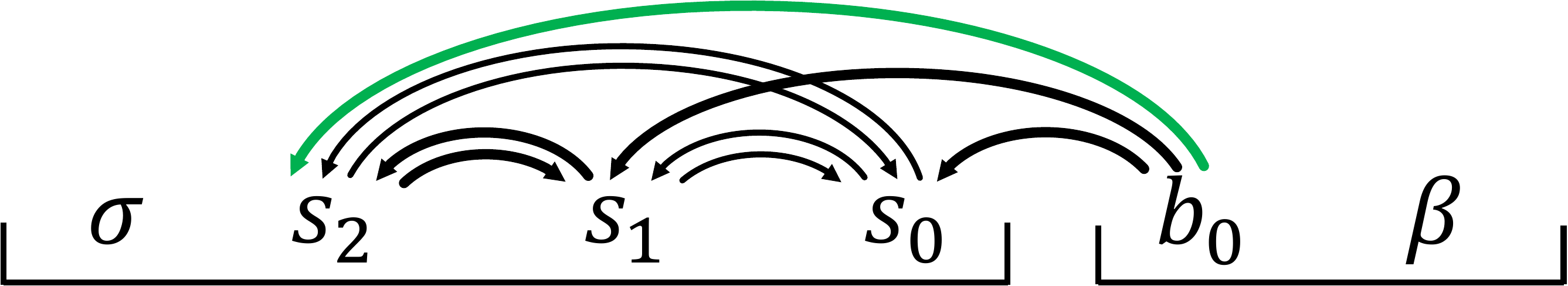}}
& $\bm{95.99\%}$ & $O(n^7)$ & $3$ \\[3pt]
Var. 3: \varthree &
\raisebox{-4pt}{
\includegraphics[width=0.4\textwidth]{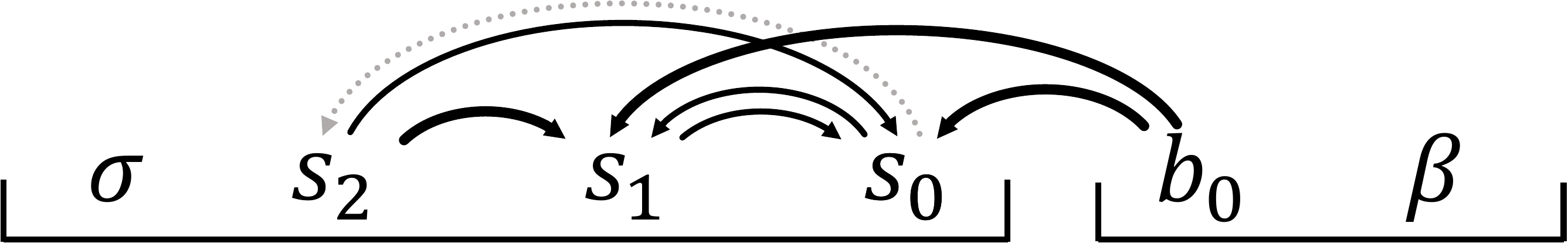}}
& $93.16\%$ & $\bm{O(n^6)}$ & $2$ \\
\bottomrule
& \hspace{45pt} stack \hfill buffer \hspace{11pt} & & & \\
\end{tabular}
\caption{\posscite{attardi} transition system of degree $2$ and
our variants.
Solid arrows
denote the inventory of reduce transitions;
each arrow points from the head to the modifier of
the edge created by that transition.
The {\em degree} of a transition is the distance between the head and modifier.
Green highlights the single degree-$3$ transition.
Thick arrows and gray dotted arrows represent additional and
deleted transitions with respect to the original
\namecite{attardi} system.
Coverage refers to the percentage of non-projective sentences
(a total of 76,084 extracted from 604,273 training sentences in UD 2.1)
that the systems are able to handle.
}
\label{fig:variants}
\end{figure*}

Non-projective dependency trees are those containing crossing edges.
They
account for
$12.59\%$ of all training sentences in the annotated Universal Dependencies (UD) 2.1 data
\cite{ud21},
and more than $20\%$  in each of 10 languages among the 54 in UD 2.1 with training
treebanks.
But
modeling non-projectivity is
computationally costly \cite{mcdonald-complexity}.

Some transition-based dependency parsers
have deduction systems that use dynamic programming
to enable
\emph{exact} inference in polynomial time and space
\cite{huang-dp,kuhlmann-dp}.
For non-projective parsing, though, the only tabularization of a transition-based parser is, to our knowledge, that of
\namecite{attardi-dp}.
They define a deduction system for
(an isomorphic variant of)
\posscite{attardi} transition system, which covers a subset of non-projective trees.
The exact inference algorithm runs in
$O(n^7)$ time,{}
 where $n$ denotes sentence length.

In this paper, we show how
\posscite{attardi-dp} system
can be modified to
generate a new family of
deduction systems with corresponding transition systems.
In particular, we present three novel variants of the degree-$2$ \citeauthor{attardi} parser,
summarized in \reffig{fig:variants}
(our technique can also be applied to generalized \namecite{attardi} systems; see \refsec{sec:variants}).
The first two
bring non-projective coverage for UD 2.1 to as high as $95.99\%$
by adding extra transitions,
and yet retain the same time complexity.
The third
reduces time complexity for exact inference to $O(n^6)$ and space complexity from $O(n^5)$ to $O(n^4)$,
while still
improving empirical coverage from $87.24\%$ to $93.16\%$.\footnote{
Faster exact inference algorithms have been defined
for some sets of mildly non-projective trees
(e.g. \newcite{pitler-1ec}; see \newcite{coverage} for more),
but
lack an underlying transition system.
Having one
has the practical advantage of allowing generative models, as in \newcite{attardi-dp},  and transition-based scoring
functions, which have
yielded good projective-parsing results \cite{exact-minfeats};
plus the theoretical advantage of
providing
a single framework supporting greedy, beam-search, and exact inference.
}
Code and full statistics for all treebanks can be found at \url{https://github.com/tzshi/nonproj-dp-variants-naacl2018}.

These theoretical improvements are a step towards making
recent state-of-the-art results in transition-based parsing with exact inference \cite{exact-minfeats} extensible to
practical non-projective parsing, by
exemplifying the  design of
transition systems with
better coverage on %
highly non-projective datasets
and, for one variant, bringing
the
runtime complexity
one level
closer to feasibility.

\section{Transition-based Parsing}
\label{sec:notation}

We first introduce necessary definitions and notation.
\subsection{A General Class of Transition Systems}
A \concept{transition system} is
given by a 4-tuple $(C, T, c^s, C_\terminal)$,
where $C$ is a set of configurations,
$T$ is a set of transition functions
between configurations, $c^s$ is an initialization function mapping an input sentence
to an initial configuration,
and $C_\terminal\subset C$ defines a set of terminal configurations.

We
employ
a tripartite representation for configurations:
$(\stack, \buffer, \arcset)$, where the three elements are as follows.
$\stack$ and $\buffer$ are disjoint lists called the \concept{stack} and \concept{buffer}, respectively.
Each dependency arc $(h, m)$ in the \concept{resolved arcs set} $A$ has head $h$ and modifier $m$.
For
a length-$n$ input sentence $w$, the initial configuration is
$c^s(w)=([], [0, 1,...,n], \emptyset)$
where the
$0$
in the initial buffer
denotes a special node representing the root of the parse tree.
All terminal configurations have an empty buffer and a stack containing only $0$.

Indexing from 0,
we write $s_i$ and $b_j$ to denote
item $i$ on the stack (starting from the right)
and
item $j$ on the buffer (from the left),
respectively.
We
use vertical bars to
separate
different parts of the buffer or stack.
For example, when concerned with the top three stack items and
the first item on the buffer,
we may write $\stack|s_2|s_1|s_0$ and $b_0|\buffer$.

\subsection{\posscite{attardi}
System}
\label{sec:attardi}

We now
introduce the
widely-used %
\namecite{attardi} system,
which includes
transitions that create arcs between non-consecutive subtrees,
thus allowing it to produce some non-projective trees.
To simplify exposition, here we present \posscite{attardi-dp}
isomorphic version.

The set of transitions
consists of a {shift} transition (\shift) and
four {reduce} transitions (\reduce).
A \concept{shift}
moves the first
buffer item onto
the stack:
\vspace{-4pt}
{\setlength{\mathindent}{0cm}
\begin{align*}
&\shift\lbrack(\sigma, b_0|\beta, A)\rbrack=(\sigma|b_0, \beta, A) .
\end{align*}
}
\noindent A
\concept{reduce} transition $\reduceargs{h}{m}$
creates a dependency arc between $h$ (head) and $m$ (modifier)
and reduces $m$. For example,
\vspace{-4pt}
{\setlength{\mathindent}{0cm}
\begin{align*}
&\reduceargs{s_0}{s_1}\lbrack(\sigma|s_1|s_0, \beta, A)\rbrack=(\sigma|s_0, \beta, A\cup\{(s_0,s_1)\})\,.
\end{align*}
}
\vspace{-1pt}
Row 1 of \reffig{fig:variants} depicts the four \citeauthor{attardi} reduces.

The distance between $h$ and $m$ in a $\reduceargs{h}{m}$ transition is called its \concept{degree}.
A system limited to degree-$1$ transitions can only parse projective sentences.
As shown in \reffig{fig:variants}, \posscite{attardi} system has two degree-2 transitions
($\reduceargs{s_0}{s_2}$ and $\reduceargs{s_2}{s_0}$)
that allow it to cover $87.24\%$ of the {\em non-projective trees} in UD 2.1.
More generally, an \citeauthor{attardi} system of degree $D$ adds $\reduceargs{s_0}{s_D}$ and $\reduceargs{s_D}{s_0}$
to
the system of degree $D-1$.

\section{Improving Coverage}

A key observation is that a degree-$D$ \citeauthor{attardi} system does not contain all possible transitions
of degree within $D$.
Since prior empirical work has ascertained that transition systems
using more transitions with degree greater than $1$ can handle more
non-projective treebank trees
\cite{attardi,coverage},
we hypothesize that
adding some of these
``missing'' reduce transitions into the system's inventory should increase coverage.
The challenge is to simultaneously maintain run-time guarantees,
as there exists a known trade-off between coverage and
complexity \cite{coverage}.
We want to  use \posscite{attardi-dp}'s exact-inference algorithm for \citeauthor{attardi}-based degree-$D$ non-projective dependency parsing systems,
which was previously analyzed as having
$O(n^{3D+1})$ time complexity.\footnote{While $O(n^7)$ or $O(n^{10})$ is not practical,
the result is still impressive, since the search space is exponential.
\namecite{attardi-dp} were inspired by \posscite{huang-dp} and \posscite{kuhlmann-dp} dynamic-programming approach for projective systems.}
{\em {\bf Our contribution} is systems that improve the \mbox{degree-2} \namecite{attardi} system's non-projective coverage,
and yet
 (i) one has degree {\bf 3}  but still the same $O(n^7) $ runtime as \namecite{attardi-dp}, rather than $O(n^{3\cdot3 + 1})$;
and (ii) another has degree 2 but {\em better} runtime than \posscite{attardi-dp} system.}
Here, we first
sketch the existing exact inference algorithm,\footnote{
See \namecite{attardi-dp} for full description.
}
and then present our variants.
\subsection{\posscite{attardi-dp} Exact Inference}
\label{sec:dp}

\begin{figure}[!t]
\centering
\includegraphics[scale=0.21]{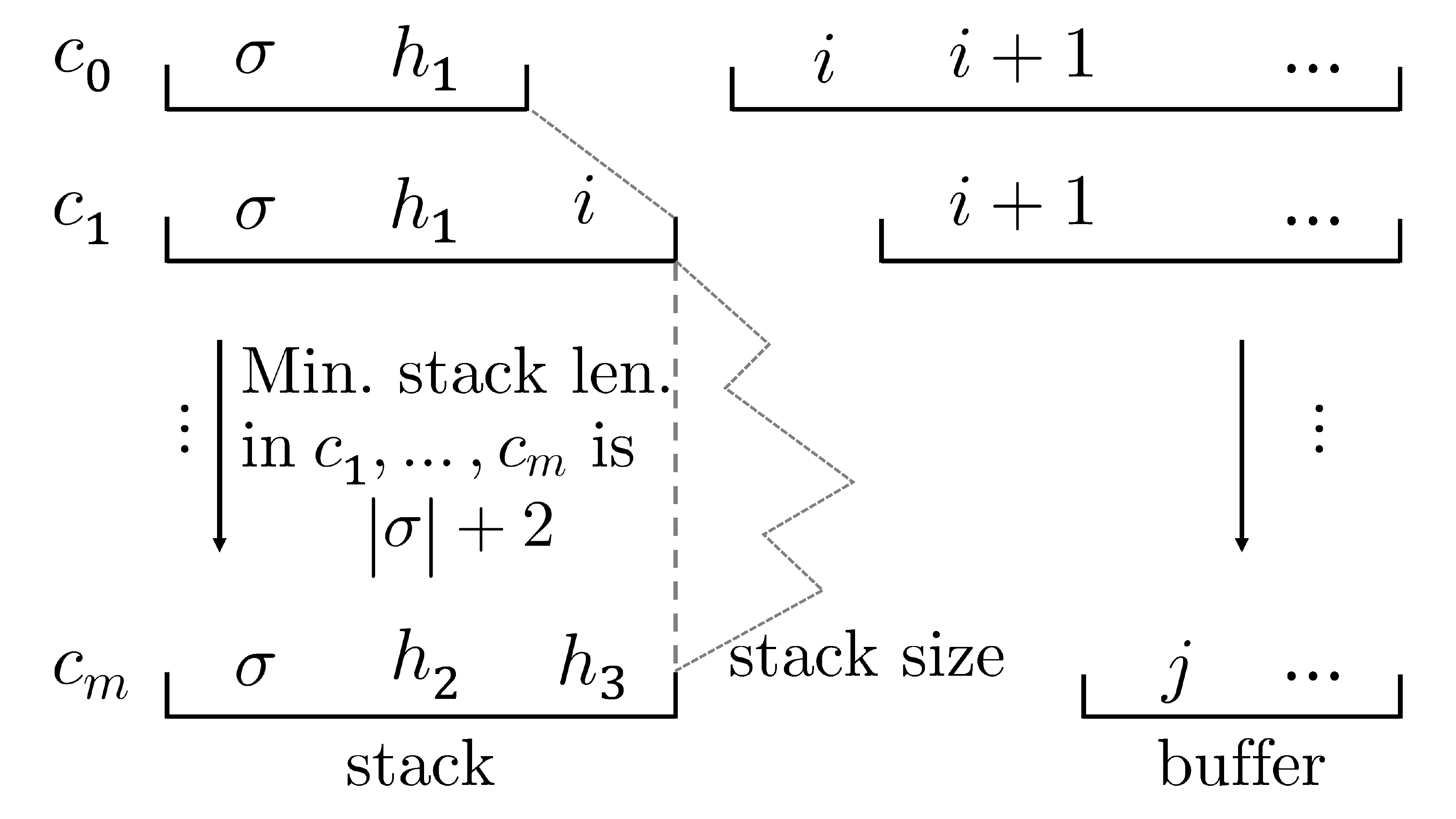}
\caption{
From \namecite[Fig.~2]{attardi-dp}:
schematic of
I-computation $[h_1, i, h_2, h_3, j]$.}
\label{fig:deduction}
\end{figure}

The main idea of the algorithm is to group transition sequences into
equivalence classes and
construct longer sequences from shorter ones.
Formally, for $m \geq 1$,  \namecite{attardi-dp} define a length-$m$ \concept{computation} as a sequence of $m$ applications of transitions to configurations:
$c_0\xrightarrow{t_1}c_1\cdots\xrightarrow{t_m}c_m$,
where $t_i\in T$ and $t_i(c_{i-1})=c_i$ for $i \in  1..m$.
As depicted in Fig.~\ref{fig:deduction}, a length-$m$ \concept{I-computation} $[h_1, i, h_2, h_3, j]$
is any length-$m$ computation where
(1) $c_0 = (\stack | h_1, i | \buffer, A)$ and $c_m = (\stack | h_2 | h_3, j | \buffer', A')$
for some $\stack$, $\buffer$, $\buffer'$, $A$, and $A'$; and
(2) for all $k \in 1..m$,  $c_k$'s stack has $\stack$ as base
and length at least $|\sigma| + 2$.
Only condition (1) is relevant to this paper:\footnote{
Condition (2) is used for proving completeness of the deduction system \cite{attardi-dp}.
}
it
states that
the net effect of an I-computation is to replace the rightmost item $h_1$ on the stack
with items $h_2$ and $h_3$, while advancing the buffer-start from $i$ to $j$.

\begin{figure}[!t]
\centering
\includegraphics[scale=0.26]{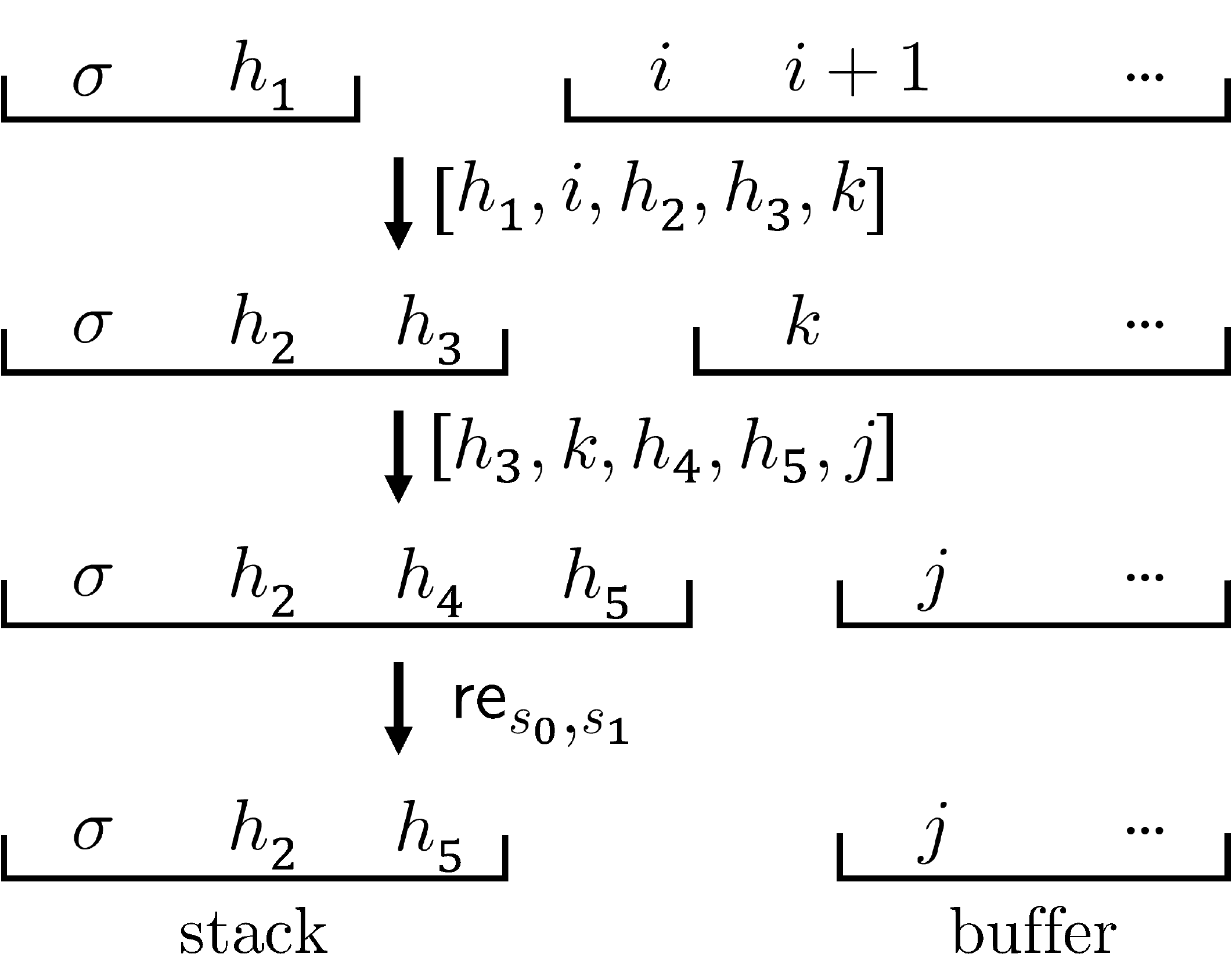}
\caption{Illustration of deduction rule $\reduceargs{s_0}{s_1}$.}
\label{fig:reduce}
\end{figure}

The dynamic programming algorithm is specified as a deduction system,
where each transition corresponds to a deduction rule.
The shift rule is:
{\setlength{\mathindent}{0cm}
\begin{align*}
&\shift:\inferrule{[h_1, i, h_2, h_3, j]}{[h_3, j, h_3, j, j+1]} .
\end{align*}
}
Each reduce rule combines two I-computations into a larger I-computation, e.g. %
(see Fig.~\ref{fig:reduce}):
{\setlength{\mathindent}{0cm}
\begin{align*}
&\reduceargs{s_0}{s_1}:\inferrule{[h_1, i, h_2, h_3, k]\quad[h_3, k, h_4, h_5, j]}{[h_1, i, h_2, h_5, j]} \,,
\end{align*}
}

\noindent
with the side condition that $h_4$ modifies $h_5.$\footnote{
This side condition can be interpreted as a grammar rule (for a recognizer) or as an edge to be scored and added to the parse tree (for a parser).
}
Other reduce transitions have similar deduction rules,
with the same two premises,
but a different conclusion depending on the reduced stack item.
As an illustration:
{\setlength{\mathindent}{0cm}
\begin{align*}
&\reduceargs{s_2}{s_0}:\inferrule{[h_1, i, h_2, h_3, k]\quad[h_3, k, h_4, h_5, j]}{[h_1, i, h_2, h_4, j]}.
\end{align*}
}

The goal of deduction is to
produce the I-computation $[\epsilon, 0, \epsilon, 0, \epsilon]$,
using the shift and reduce deduction rules
starting from the axiom $[\epsilon, 0, \epsilon, 0, 1]$, corresponding to the first and mandatory shift transition moving the root node from buffer to stack.
$\epsilon$ stands for an empty stack or buffer.
As analyzed by \citet{attardi-dp},
direct tabularization for this deduction system takes $O(n^5)$ space and $O(n^8)$ time.
With adaptation of the ``hook trick'' described in \citet{eisner+satta:99},
we can reduce the running time to $O(n^7)$.

\subsection{Our New Variants}
\label{sec:variants}

In this section,
we modify  \posscite{attardi-dp} set of reduce deduction rules
to improve coverage or time complexity.
Since each
such deduction rule corresponds to a reduce transition,
each revision to the deduction system
yields a variant of  \posscite{attardi} parser.
In other words, generalization of the deduction system gives rise to a family of non-projective transition-based dependency parsers.

We first explain why there are exactly nine reduce transitions $\ruleset = \{
\reduceargs{s_0}{s_1},\allowbreak
\reduceargs{s_1}{s_0},\allowbreak
\reduceargs{s_0}{s_2},\allowbreak
\reduceargs{s_2}{s_0},\allowbreak
\reduceargs{s_1}{s_2},\allowbreak
\reduceargs{s_2}{s_1},\allowbreak
\reduceargs{b_0}{s_0},\allowbreak
\reduceargs{b_0}{s_1},\allowbreak
\reduceargs{b_0}{s_2}
\}$ that can be used in \posscite{attardi-dp} exact inference algorithm,
without allowing a reduction with head $b_i$ for $i \geq 1$.\footnote{Such reductions might prove interesting
in the future.} (Note that \posscite{attardi-dp} reduce rules are precisely the first four
elements of $\ruleset$.)
From \reffig{fig:reduce} we infer that the concatenation of  I-computations  $[h_1, i, h_2, h_3, k]$ and $[h_3, k, h_4,
h_5, j]$ yields a
configuration of the form $(\stack|h_2|h_4|h_5, j|\buffer, A)$. For the application of a reduce rule to yield a valid
I-computation, by condition (1) of the I-computation definition, first, the head and modifier must be selected from the
``exposed'' elements $h_2$, $h_4$, $h_5$, and
$j$, corresponding to $s_2$, $s_1$, $s_0$, $b_0$, respectively; and second, the modifier can only come from the stack. $\ruleset$ is precisely the set of rules satisfying these
criteria.
Further, every reduce transition from $\ruleset$ is compatible with \posscite{eisner+satta:99} ``hook trick''.
This gives us the satisfactory result that the $O(n^7)$ running time upper bound still holds for transitions in $\ruleset$, even though one of them has degree 3.

Next, we consider three notable variants within the family of
$\ruleset$-based
 non-projective transition-based dependency parsers.
They are given
in \reffig{fig:variants},
along with their time complexities and empirical coverage statistics.
The latter is computed using static oracles \cite{oracle}
on the UD 2.1 dataset \cite{ud21}.\footnote{
We also compare results from symbolic execution of the dynamic programming algorithms on short sentences
as a double check.
}
We report the global coverage over the
76,084 non-projective sentences from all the training treebanks.
One might assume that adding more degree-1 transitions wouldn't improve coverage of trees with non-crossing edges.
On the other hand, since their addition doesn't affect the asymptotic run-time, we define
{\bf \varone} to include all five degree-1 transitions from $\ruleset$ into the \namecite{attardi} system.
Surprisingly, using \varone improves
non-projective coverage from $87.24\%$ to $93.32\%$.
Furthermore, recall that we argued above that, by construction, using any of the transitions in
$\ruleset$ still preserves the original  $O(n^7)$ run-time
upper bound for \posscite{attardi-dp} exact inference
algorithm.
We therefore define
{\bf \vartwo} to include all 9 reduce transitions in $\ruleset$; it runs in time $O(n^7)$  despite
the fact that $\reduceargs{b_0}{s_2} \in \ruleset$ has degree 3, a significant improvement over the best
previously-known bound for degree-3 systems of $O(n^{10})$.
Moreover,  as shown in \reffig{fig:variants}, this variant improves non-projective coverage to $95.54\%$.
Now, if our goal is to reduce run-time, we can start with an \namecite{attardi} system of degree 1 instead of 2, which,
as previously mentioned, can only handle projective sentences, but which has runtime $O(n^{(3\cdot1)+1})= O(n^4)$.
Reasoning about the analog of $\ruleset$ with respect to \posscite{kuhlmann-dp} exact inference algorithm --- the
projective predecessor of \namecite{attardi-dp} --- brings us to the degree-2 set of reduce rules
$\{
\reduceargs{s_0}{s_1},\allowbreak
\reduceargs{s_1}{s_0},\allowbreak
\reduceargs{b_0}{s_1}\allowbreak
\}$.
This system, however, can only handle leftward non-projective arcs.
Instead, we return to \vartwo, but discard transitions reducing $s_2$, thus deriving
{\bf \varthree},
which still
produces both left and right
non-projective arcs, but
has a run-time lower than $O(n^7)$, which we show as follows.
Since $s_2$ cannot be reduced,
when concatenating $[h_1, i, h_2, h_3, k]$ and $[h_3, k, h_4, h_5, j]$,
the larger I-computation we deduce will be either
$[h_1, i, h_2, h_4, j]$ or $[h_1, i, h_2, h_5, j]$,
so that
the first three indices of the conclusion item remain the same as those of the first premise.
In addition,
the only remaining deduction rule, a shift,
produces deduction items
of the form $[h_1, j, h_1, j, j+1]$.
Hence, all derivable items will be
of the form $[h_1, i, h_1, h_3, j]$,
with only four unique indices, instead of five.
It follows that the exact inference algorithm for this variant runs in $O(n^6)$ time, improving from $O(n^7)$.
The tabularization takes $O(n^4)$ space, a reduction from
the
original $O(n^5)$ as well.
In terms of empirical coverage,
this new system can handle $93.16\%$ of the non-projective sentences
in UD 2.1,
more than \posscite{attardi} system,
but fewer than
our other two variants.

Generally, for a {\em degree}-$D$ \namecite{attardi}-based system,
one may apply our first two variants to improve its non-projective coverage
while maintaining the previously-analyzed $O(n^{3D+1})$ time complexity,
or the third variant to reduce its time complexity down to $O(n^{3D})$,
and space complexity from $O(n^{2D+1})$ to $O(n^{2D})$.

\section{Conclusion}
\label{sec:conclusion}

We have introduced a family of variants of
\posscite{attardi-dp} \citeauthor{attardi}-based transition system
and its associated dynamic programming algorithm.
Among
these, we have highlighted novel algorithms that
(1) increase non-projective coverage without affecting computational complexity for exact inference,
and (2) improve the time and space complexity for exact inference,
even while providing
better
coverage than the original parser.
Specifically, our \varthree runs in $O(n^6)$ time and $O(n^4)$ space (improving from $O(n^7)$ and $O(n^5)$, respectively)
while providing coverage
of 93.16\% of the non-projective sentences in UD 2.1.

Exact inference for transition-based parsers has recently achieved state-of-the-art results in projective parsing
\cite{exact-minfeats}.
The complexity improvements achieved in this paper are a step towards making
their exact-inference,
projective
approach
extensible to practical, wide-coverage non-projective parsing.

\section*{Acknowledgments}
\label{sec:acknowledgments}

We thank the reviewers for their time and careful reading,
and reviewer 2 for suggestions that improved our presentation.
TS and LL were supported in part by a Google Focused Research Grant to Cornell University.
CG
has received funding from the European
Research Council (ERC), under the European
Union's Horizon 2020 research and innovation
programme (FASTPARSE, grant agreement No
714150), from the TELEPARES-UDC project
(FFI2014-51978-C2-2-R) and the ANSWER-ASAP project (TIN2017-85160-C2-1-R) from MINECO, and from Xunta de Galicia (ED431B 2017/01).
LL was also supported in part by NSF grant SES-1741441. Any
opinions, findings, and conclusions or recommendations expressed in
this material are those of the author(s) and do not necessarily
reflect the views of the National Science Foundation or other
sponsors.

\bibliography{ref}

\begin{thebibliography}{11}
\expandafter\ifx\csname natexlab\endcsname\relax\def\natexlab#1{#1}\fi

\bibitem[{Attardi(2006)}]{attardi}
Giuseppe Attardi. 2006.
\newblock \href {http://www.aclweb.org/anthology/W/W06/W06-2922} {Experiments
  with a multilanguage non-projective dependency parser}.
\newblock In \emph{Proceedings of the Tenth Conference on Computational Natural
  Language Learning (CoNLL-X)}, pages 166--170, New York City, New York, USA.

\bibitem[{Cohen et~al.(2011)Cohen, G\'{o}mez-Rodr\'{i}guez, and
  Satta}]{attardi-dp}
Shay~B. Cohen, Carlos G\'{o}mez-Rodr\'{i}guez, and Giorgio Satta. 2011.
\newblock \href {http://www.aclweb.org/anthology/D11-1114} {Exact inference for
  generative probabilistic non-projective dependency parsing}.
\newblock In \emph{Proceedings of the 2011 Conference on Empirical Methods in
  Natural Language Processing (EMNLP)}, pages 1234--1245, Edinburgh, Scotland,
  UK.

\bibitem[{Cohen et~al.(2012)Cohen, G{\'{o}}mez{-}Rodr{\'{\i}}guez, and
  Satta}]{oracle}
Shay~B. Cohen, Carlos G{\'{o}}mez{-}Rodr{\'{\i}}guez, and Giorgio Satta. 2012.
\newblock \href {http://arxiv.org/abs/1206.6735} {Elimination of spurious
  ambiguity in transition-based dependency parsing}.
\newblock \emph{ArXiv e-prints}, 1206.6735.

\bibitem[{Eisner and Satta(1999)}]{eisner+satta:99}
Jason Eisner and Giorgio Satta. 1999.
\newblock \href {http://aclweb.org/anthology/P99-1059} {Efficient parsing for
  bilexical context-free grammars and head automaton grammars}.
\newblock In \emph{Proceedings of the 37th Annual Meeting of the Association
  for Computational Linguistics (ACL)}, pages 457--464, College Park, Maryland,
  USA.

\bibitem[{G\'{o}mez-Rodr\'{i}guez(2016)}]{coverage}
Carlos G\'{o}mez-Rodr\'{i}guez. 2016.
\newblock \href {https://doi.org/10.1162/COLI_a_00267} {Restricted
  non-projectivity: Coverage vs. efficiency}.
\newblock \emph{Computational Linguistics}, 42(4):809--817.

\bibitem[{Huang and Sagae(2010)}]{huang-dp}
Liang Huang and Kenji Sagae. 2010.
\newblock \href {http://www.aclweb.org/anthology/P10-1110} {Dynamic programming
  for linear-time incremental parsing}.
\newblock In \emph{Proceedings of the 48th Annual Meeting of the Association
  for Computational Linguistics (ACL)}, pages 1077--1086, Uppsala, Sweden.

\bibitem[{Kuhlmann et~al.(2011)Kuhlmann, G\'{o}mez-Rodr\'{i}guez, and
  Satta}]{kuhlmann-dp}
Marco Kuhlmann, Carlos G\'{o}mez-Rodr\'{i}guez, and Giorgio Satta. 2011.
\newblock \href {http://www.aclweb.org/anthology/P11-1068} {Dynamic programming
  algorithms for transition-based dependency parsers}.
\newblock In \emph{Proceedings of the 49th Annual Meeting of the Association
  for Computational Linguistics: Human Language Technologies (ACL-HLT)}, pages
  673--682, Portland, Oregon, USA.

\bibitem[{McDonald and Satta(2007)}]{mcdonald-complexity}
Ryan McDonald and Giorgio Satta. 2007.
\newblock \href {http://www.aclweb.org/anthology/W/W07/W07-2216} {On the
  complexity of non-projective data-driven dependency parsing}.
\newblock In \emph{Proceedings of the Tenth International Conference on Parsing
  Technologies (IWPT)}, pages 121--132, Prague, Czech Republic. Association for
  Computational Linguistics.

\bibitem[{Nivre et~al.(2017)Nivre, Agi{\'c}, Ahrenberg, Antonsen, Aranzabe,
  Asahara, Ateyah, Attia, Atutxa, Augustinus, Badmaeva, Ballesteros, Banerjee,
  Bank, Barbu~Mititelu, Bauer, Bengoetxea, Bhat, Bick, Bobicev, B{\"o}rstell,
  Bosco, Bouma, Bowman, Burchardt, Candito, Caron, Cebiro{\u g}lu~Eryi{\u g}it,
  Celano, Cetin, Chalub, Choi, Cinkov{\'a}, {\c C}{\"o}ltekin, Connor,
  Davidson, de~Marneffe, de~Paiva, Diaz~de Ilarraza, Dirix, Dobrovoljc, Dozat,
  Droganova, Dwivedi, Eli, Elkahky, Erjavec, Farkas, Fernandez~Alcalde, Foster,
  Freitas, Gajdo{\v s}ov{\'a}, Galbraith, Garcia, G{\"a}rdenfors, Gerdes,
  Ginter, Goenaga, Gojenola, G{\"o}k{\i}rmak, Goldberg, G{\'o}mez~Guinovart,
  Gonz{\'a}les~Saavedra, Grioni, Gr{\=u}z{\={\i}}tis, Guillaume, Habash,
  Haji{\v c}, Haji{\v c}~jr., H{\`a}~M{\~y}, Harris, Haug, Hladk{\'a},
  Hlav{\'a}{\v c}ov{\'a}, Hociung, Hohle, Ion, Irimia, Jel{\'{\i}}nek,
  Johannsen, J{\o}rgensen, Ka{\c s}{\i}kara, Kanayama, Kanerva, Kayadelen,
  Kettnerov{\'a}, Kirchner, Kotsyba, Krek, Laippala, Lambertino, Lando, Lee,
  L{\^e}~H{\`{\^o}}ng, Lenci, Lertpradit, Leung, Li, Li, Li, Ljube{\v s}i{\'c},
  Loginova, Lyashevskaya, Lynn, Macketanz, Makazhanov, Mandl, Manning, M{\u
  a}r{\u a}nduc, Mare{\v c}ek, Marheinecke, Mart{\'{\i}}nez~Alonso, Martins,
  Ma{\v s}ek, Matsumoto, {McDonald}, Mendon{\c c}a, Miekka, Missil{\"a},
  Mititelu, Miyao, Montemagni, More, Moreno~Romero, Mori, Moskalevskyi,
  Muischnek, M{\"u}{\"u}risep, Nainwani, Nedoluzhko, Ne{\v
  s}pore-B{\=e}rzkalne, Nguy{\~{\^e}}n~Th{\d i}, Nguy{\~{\^e}}n Th{\d i}~Minh,
  Nikolaev, Nurmi, Ojala, Osenova, {\"O}stling, {\O}vrelid, Pascual,
  Passarotti, Perez, Perrier, Petrov, Piitulainen, Pitler, Plank, Popel,
  Pretkalni{\c n}a, Prokopidis, Puolakainen, Pyysalo, Rademaker, Ramasamy,
  Rama, Ravishankar, Real, Reddy, Rehm, Rinaldi, Rituma, Romanenko, Rosa,
  Rovati, Sagot, Saleh, Samard{\v z}i{\'c}, Sanguinetti, Saul{\={\i}}te,
  Schuster, Seddah, Seeker, Seraji, Shen, Shimada, Sichinava, Silveira, Simi,
  Simionescu, Simk{\'o}, {\v S}imkov{\'a}, Simov, Smith, Stella, Straka,
  Strnadov{\'a}, Suhr, Sulubacak, Sz{\'a}nt{\'o}, Taji, Tanaka, Trosterud,
  Trukhina, Tsarfaty, Tyers, Uematsu, Ure{\v s}ov{\'a}, Uria, Uszkoreit,
  Vajjala, van Niekerk, van Noord, Varga, Villemonte de~la Clergerie, Vincze,
  Wallin, Washington, Wir{\'e}n, Wong, Yu, {\v Z}abokrtsk{\'y}, Zeldes, Zeman,
  and Zhu}]{ud21}
Joakim Nivre, {\v Z}eljko Agi{\'c}, Lars Ahrenberg, Lene Antonsen, Maria~Jesus
  Aranzabe, Masayuki Asahara, Luma Ateyah, Mohammed Attia, Aitziber Atutxa,
  Liesbeth Augustinus, Elena Badmaeva, Miguel Ballesteros, Esha Banerjee,
  Sebastian Bank, Verginica Barbu~Mititelu, John Bauer, Kepa Bengoetxea,
  Riyaz~Ahmad Bhat, Eckhard Bick, Victoria Bobicev, Carl B{\"o}rstell, Cristina
  Bosco, Gosse Bouma, Sam Bowman, Aljoscha Burchardt, Marie Candito, Gauthier
  Caron, G{\"u}l{\c s}en Cebiro{\u g}lu~Eryi{\u g}it, Giuseppe G.~A. Celano,
  Savas Cetin, Fabricio Chalub, Jinho Choi, Silvie Cinkov{\'a}, {\c C}a{\u
  g}r{\i} {\c C}{\"o}ltekin, Miriam Connor, Elizabeth Davidson, Marie-Catherine
  de~Marneffe, Valeria de~Paiva, Arantza Diaz~de Ilarraza, Peter Dirix, Kaja
  Dobrovoljc, Timothy Dozat, Kira Droganova, Puneet Dwivedi, Marhaba Eli, Ali
  Elkahky, Toma{\v z} Erjavec, Rich{\'a}rd Farkas, Hector Fernandez~Alcalde,
  Jennifer Foster, Cl{\'a}udia Freitas, Katar{\'{\i}}na Gajdo{\v s}ov{\'a},
  Daniel Galbraith, Marcos Garcia, Moa G{\"a}rdenfors, Kim Gerdes, Filip
  Ginter, Iakes Goenaga, Koldo Gojenola, Memduh G{\"o}k{\i}rmak, Yoav Goldberg,
  Xavier G{\'o}mez~Guinovart, Berta Gonz{\'a}les~Saavedra, Matias Grioni,
  Normunds Gr{\=u}z{\={\i}}tis, Bruno Guillaume, Nizar Habash, Jan Haji{\v c},
  Jan Haji{\v c}~jr., Linh H{\`a}~M{\~y}, Kim Harris, Dag Haug, Barbora
  Hladk{\'a}, Jaroslava Hlav{\'a}{\v c}ov{\'a}, Florinel Hociung, Petter Hohle,
  Radu Ion, Elena Irimia, Tom{\'a}{\v s} Jel{\'{\i}}nek, Anders Johannsen,
  Fredrik J{\o}rgensen, H{\"u}ner Ka{\c s}{\i}kara, Hiroshi Kanayama, Jenna
  Kanerva, Tolga Kayadelen, V{\'a}clava Kettnerov{\'a}, Jesse Kirchner, Natalia
  Kotsyba, Simon Krek, Veronika Laippala, Lorenzo Lambertino, Tatiana Lando,
  John Lee, Ph{\uhorn}{\ohorn}ng L{\^e}~H{\`{\^o}}ng, Alessandro Lenci, Saran Lertpradit,
  Herman Leung, Cheuk~Ying Li, Josie Li, Keying Li, Nikola Ljube{\v s}i{\'c},
  Olga Loginova, Olga Lyashevskaya, Teresa Lynn, Vivien Macketanz, Aibek
  Makazhanov, Michael Mandl, Christopher Manning, C{\u a}t{\u a}lina M{\u
  a}r{\u a}nduc, David Mare{\v c}ek, Katrin Marheinecke, H{\'e}ctor
  Mart{\'{\i}}nez~Alonso, Andr{\'e} Martins, Jan Ma{\v s}ek, Yuji Matsumoto,
  Ryan {McDonald}, Gustavo Mendon{\c c}a, Niko Miekka, Anna Missil{\"a}, C{\u
  a}t{\u a}lin Mititelu, Yusuke Miyao, Simonetta Montemagni, Amir More, Laura
  Moreno~Romero, Shinsuke Mori, Bohdan Moskalevskyi, Kadri Muischnek, Kaili
  M{\"u}{\"u}risep, Pinkey Nainwani, Anna Nedoluzhko, Gunta Ne{\v
  s}pore-B{\=e}rzkalne, L{\uhorn}{\ohorn}ng Nguy{\~{\^e}}n~Th{\d i}, Huy{\`{\^e}}n
  Nguy{\~{\^e}}n Th{\d i}~Minh, Vitaly Nikolaev, Hanna Nurmi, Stina Ojala,
  Petya Osenova, Robert {\"O}stling, Lilja {\O}vrelid, Elena Pascual, Marco
  Passarotti, Cenel-Augusto Perez, Guy Perrier, Slav Petrov, Jussi Piitulainen,
  Emily Pitler, Barbara Plank, Martin Popel, Lauma Pretkalni{\c n}a, Prokopis
  Prokopidis, Tiina Puolakainen, Sampo Pyysalo, Alexandre Rademaker, Loganathan
  Ramasamy, Taraka Rama, Vinit Ravishankar, Livy Real, Siva Reddy, Georg Rehm,
  Larissa Rinaldi, Laura Rituma, Mykhailo Romanenko, Rudolf Rosa, Davide
  Rovati, Beno{\^{\i}}t Sagot, Shadi Saleh, Tanja Samard{\v z}i{\'c}, Manuela
  Sanguinetti, Baiba Saul{\={\i}}te, Sebastian Schuster, Djam{\'e} Seddah,
  Wolfgang Seeker, Mojgan Seraji, Mo~Shen, Atsuko Shimada, Dmitry Sichinava,
  Natalia Silveira, Maria Simi, Radu Simionescu, Katalin Simk{\'o}, M{\'a}ria
  {\v S}imkov{\'a}, Kiril Simov, Aaron Smith, Antonio Stella, Milan Straka,
  Jana Strnadov{\'a}, Alane Suhr, Umut Sulubacak, Zsolt Sz{\'a}nt{\'o}, Dima
  Taji, Takaaki Tanaka, Trond Trosterud, Anna Trukhina, Reut Tsarfaty, Francis
  Tyers, Sumire Uematsu, Zde{\v n}ka Ure{\v s}ov{\'a}, Larraitz Uria, Hans
  Uszkoreit, Sowmya Vajjala, Daniel van Niekerk, Gertjan van Noord, Viktor
  Varga, Eric Villemonte de~la Clergerie, Veronika Vincze, Lars Wallin,
  Jonathan~North Washington, Mats Wir{\'e}n, Tak-sum Wong, Zhuoran Yu, Zden{\v
  e}k {\v Z}abokrtsk{\'y}, Amir Zeldes, Daniel Zeman, and Hanzhi Zhu. 2017.
\newblock \href {http://hdl.handle.net/11234/1-2515} {{U}niversal
  {D}ependencies 2.1}.
\newblock {LINDAT}/{CLARIN} digital library at the Institute of Formal and
  Applied Linguistics ({{\'U}FAL}), Faculty of Mathematics and Physics, Charles
  University.

\bibitem[{Pitler et~al.(2013)Pitler, Kannan, and Marcus}]{pitler-1ec}
Emily Pitler, Sampath Kannan, and Mitchell Marcus. 2013.
\newblock \href {http://www.aclweb.org/anthology/Q13-1002} {Finding optimal
  1-endpoint-crossing trees}.
\newblock \emph{Transactions of the Association of Computational Linguistics},
  1:13--24.

\bibitem[{Shi et~al.(2017)Shi, Huang, and Lee}]{exact-minfeats}
Tianze Shi, Liang Huang, and Lillian Lee. 2017.
\newblock \href {https://www.aclweb.org/anthology/D17-1002} {Fast(er) exact
  decoding and global training for transition-based dependency parsing via a
  minimal feature set}.
\newblock In \emph{Proceedings of the 2017 Conference on Empirical Methods in
  Natural Language Processing (EMNLP)}, pages 12--23, Copenhagen, Denmark.

\end{thebibliography}
\bibliographystyle{acl_natbib}

\end{document}